\documentclass{article}
\usepackage{spconf,amsmath,epsfig}

\usepackage{algorithm}
\usepackage{algorithmic}
\usepackage{amsmath,bm,bbm}
\usepackage{multirow}
\usepackage{graphicx}
\usepackage{makecell}

\usepackage{color}
\usepackage{xcolor}
\usepackage{colortbl}

\usepackage{pifont}
\newcommand{\etal}{\textit{et al}.}
\newcommand{\ie}{\textit{i}.\textit{e}.}
\newcommand{\eg}{\textit{e}.\textit{g}.}

\newcommand{\etc}{\textit{etc}}

\usepackage[pagebackref,breaklinks,colorlinks]{hyperref}

\renewcommand{\paragraph}[1]{\noindent\textbf{#1}\quad}

\let\OLDthebibliography\thebibliography
\renewcommand\thebibliography[1]{
  \OLDthebibliography{#1}
  \setlength{\parskip}{0pt}
  \setlength{\itemsep}{0pt plus 0.3ex}
}

\begin{document}\sloppy

\def\x{{\mathbf x}}
\def\L{{\cal L}}

\title{Convex Combination Consistency between Neighbors for Weakly-supervised Action Localization}
%
\name{Qinying Liu, Zilei Wang, Ruoxi Chen, and Zhilin Li}
\address{\{lydyc,ruoxi16,lizhilin\}@mail.ustc.edu.cn; zlwang@ustc.edu.cn}

\maketitle

\begin{abstract}
Weakly-supervised temporal action localization (WTAL) intends to detect action instances with only weak supervision, \eg, video-level labels. The current~\textit{de facto} pipeline locates action instances by thresholding and grouping continuous high-score regions on temporal class activation sequences. In this route, the capacity of the model to recognize the relationships between adjacent snippets is of vital importance which determines the quality of the action boundaries. However, it is error-prone since the variations between adjacent snippets are typically subtle, and unfortunately this is overlooked in the literature. To tackle the issue, we propose a novel WTAL approach named Convex Combination Consistency between Neighbors (C$^3$BN).  C$^3$BN consists of two key ingredients: a micro data augmentation strategy that increases the diversity in-between adjacent snippets by convex combination of adjacent snippets, and a macro-micro consistency regularization that enforces the model to be invariant to the transformations~\textit{w.r.t.} video semantics, snippet predictions, and snippet representations. Consequently, fine-grained patterns in-between adjacent snippets are enforced to be explored, thereby resulting in a more robust action boundary localization. Experimental results demonstrate the effectiveness of C$^3$BN on top of various baselines for WTAL with video-level and point-level supervisions. Code is at \href{https://github.com/Qinying-Liu/C3BN}{C3BN}.
\end{abstract}

\section{Introduction}
\label{sec:intro}
Temporal action localization~\cite{SSN,liu2023improve,liu2020progressive} intends to localize action instances and recognize their categories in videos. In recent years, numerous works delve into the fully supervised TAL and gain significant improvement. However, these methods require tremendous manual frame-level annotations, which is expensive and time-consuming. 
Recently, weakly-supervised TAL (WTAL)\cite{UntrimmedNet,liu2023revisiting} has received increasing attention, as it allows us to detect the action instances with only weak supervision, \eg, video-level labels~\cite{UntrimmedNet} and point-level labels~\cite{Moltisanti2019CVPR,yang2021background}. 
In particular, video-level labels are the most commonly used.

\begin{figure}[t]
\centering
\vspace{-4mm}
\includegraphics[width=0.65\linewidth]{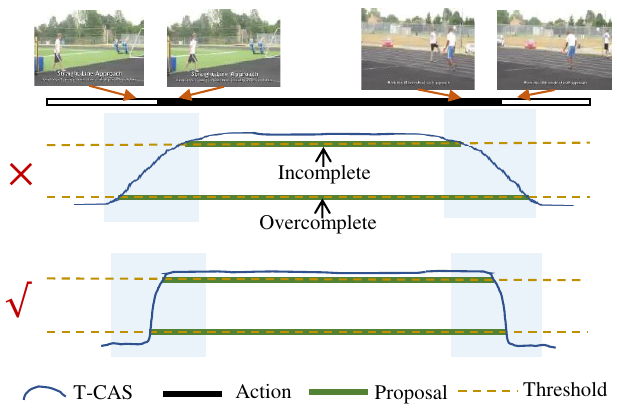}
\centering
\vspace{-3mm}
\caption{\textbf{Motivation of our method.} Due to the vague distinctions between adjacent snippets, an under-performing model may produce similar activations for these snippets, resulting in incomplete/overcomplete proposals (see the upper T-CA). In this paper, we expect the model to correctly classify adjacent snippets, thereby localizing accurate boundaries  (see the lower T-CAS). }
\vspace{-3mm}
\label{fig:issue}
\end{figure}

Mainstream WTAL methods~\cite{UntrimmedNet,ma2020sfnet}, regardless of the types of weak supervisions, employ a video action classification model to learn the Temporal Class Activation Sequence (T-CAS). 
After training, they utilize the T-CAS to localize action in a bottom-up fashion~\cite{SSN} derived from the watershed algorithm. 
Specifically, it consists of two main steps. \ding{182} \textit{boundary localization}: generating action proposals by thresholding and merging the continuous action regions of T-CAS with multiple thresholds; \ding{183} \textit{proposal evaluation}: calculating proposal-level scores by aggregating snippet-level scores within the regions. 
Recent methods pay many efforts to learn accurate snippet-level scores by various techniques, \eg, pseudo-labeling~\cite{huang2022weakly,Yang_2021_UGCT} and contrastive learning~\cite{liexploring,CoLA}. In other words, these methods focus on the semantic relationships between each snippet and global class centers/other snippets. Despite the progress, we argue that they may be sub-optimal since what really matters in \ding{182} is the relationship between adjacent snippets~\cite{kang2022uboco}. 
As depicted in Fig.~\ref{fig:issue}, adjacent snippets are usually similar in content and thus have close activation, which may cause incomplete or over-complete proposals.
Hence, it is necessary to enable the model to be sensitive enough to the fine-grained distinctions between adjacent snippets.

To counteract this issue, we introduce a plug-and-play training strategy dubbed Convex Combination Consistency Between Neighbors (C$^3$BN) for WTAL. 
The idea of our work stems from MixUp~\cite{zhang2018mixup}, where the classification model trained on the mixture of image pairs achieves promising performance. In light of this, to enhance the ability of the WTAL model to distinguish adjacent snippets, we propose a micro~\footnote{By ‘micro’, we mean the proposed data augmentation strategy is on snippets rather than videos.} data augmentation strategy, where the pairs of adjacent snippets (termed as~\textit{parent snippets}) are mixed by convex combination to generate a set of new snippets (termed as~\textit{child snippets}).   
However, there are still two problems that need to be handled before using the child snippets. The first problem is how to feed the child snippets into WTAL models. Unlike conventional MixUp which treats images as independent instances, most WTAL models take the snippet sequences as input, followed by a few temporal convolution layers to enlarge the temporal receptive field. Therefore, we have to define the temporal orders of the child snippets before they can be processed by the models. To address the challenge, we propose to take advantage of the \textit{temporal continuity} prior in videos~\cite{kalogeiton2017action}: the scenes usually change smoothly and continuously along temporal dimension. This property implies that the temporal location of each child snippet lies in-between that of its parent snippets. With the virtual locations, we arrange the child snippets of a video into a new sequence, which can be viewed as a locally deformed version of the original sequence.

The second problem is how to utilize the child snippets to promote model training. 
In MixUp, the mixed sample is assigned with the mixture of the ground-truth labels of the original samples, encouraging the model to behave linearly in-between samples. 
In our case, however, only weak labels of snippets are available. To this end, we develop a macro-micro consistency regularization, which makes use of both weak supervision and linear behaviour to regularize the model training . 
Specifically, we introduce three consistency regularization terms to exploit different relationships between child and parent snippets~\textit{w.r.t.} video semantics, snippet predictions and snippet representations, thereby facilitating model training from~\textit{macro view} to~\textit{micro view} and from~\textit{low-level representations} to~\textit{high-level semantics}. In this way, more fine-grained cues in-between adjacent snippets are preserved, eventually improving the robustness of boundary localization. 

The idea behind C$^{3}$BN is generic and conceptually complementary to other methods, which is justified by the performance promotion on a variety of base approaches and datasets. More importantly, extensive quantitative and qualitative results verify the efficacy of C$^3$BN in \ding{182} boundary localization. Hence, our contributions are:
1) We propose to consider the potential of adjacent snippets in WTAL and then design a micro data augmentation strategy by convex combination of adjacent snippets. 2) We propose three regularization terms to enhance the consistency properties \textit{w.r.t.} video semantics, snippet predictions and snippet features. 
3) Our method can be easily plugged into existing WTAL methods with either video-level supervision or point-level supervision.


\vspace{-2mm}
\section{Related Work} \label{sec:related_work}

\noindent\textbf{Data Augmentation} aims to enlarge the train set using transformations. Conventional image transformations include cropping, flips, rotation,~\etc. Recent studies consider employing multiple images for augmentation. MixUp~\cite{zhang2018mixup} proposes to combine the pixel values and labels of two images by linear interpolation. It has been proven effective for the classification task, which is followed by~\cite{verma2019manifold}. Our method employ the idea of instance mixtures with task-specific designs. Concretely, we achieve the mixture operation on two snippets within a video rather than on two different videos likewise MixUp, making the perturbations to snippets more controllable for incorporating the proposed method into the existing WTAL frameworks. 

\noindent\textbf{MixUp} trains a model by linearly
interpolating two training examples and their labels~\cite{zhang2018mixup}.
It is proven effective for the classification task, followed by different variants. For instance,~\cite{verma2019manifold} extends the linear interpolation from input-level to feature-level. Recently, extensive methods are proposed to incorporate MixUp with semantic segmentation~\cite{chang2020mixup}, self-supervised learning~\cite{kim2020mixco}, \etc. Our method also employs the idea of instance mixtures, but it is a not trivial extension of previous methods.     
For example, the original MixUp mixes two randomly selected images. Extending it directly from image to snippet will cause the locations of the generated snippets undefined. In addition, according to~\cite{chang2020mixup}, another alternative is to mix two random videos snippet-by-snippet. It is also not feasible for WTAL as the video lengths of two videos fed into the models may be different~\cite{huang2021foreground} in practice. ~\textit{Different from above methods, we achieve the mixture operation on two adjacent snippets within a video, yielding more controllable perturbation to snippets for incorporating the method into the existing WTAL models.}       

\noindent\textbf{Consistency regularization} is a crucial technique in semi-supervised learning. It is assumed that a classifier should output the same class probability for an unlabeled sample even after it is augmented. Prior works~\cite{temporal,consistency} apply the consistency regularization on different augmentations of an unlabeled sample. After that, several variants~\cite{mixmatch,fixmatch} are further proposed to extend its applications. Among them, MixMatch~\cite{mixmatch} also uses MixUp by mixing unlabeled samples and their pseudo-labels.The differences between our method and them are~\textit{1) MixMatch randomly mixes two examples, while we only mix the adjacent snippets; 2) MixMatch guesses the hard pseudo-labels of unlabeled samples and relies on a complicated ensemble of multiple predictions to improve the quality of pseudo-labels, whereas we do not guess the hard pseudo-labels of unlabeled samples, thereby reducing undesirable label noise~\cite{tang2021humble,liu2022collaborating}. }

\noindent\textbf{Self-supervised contrastive learning}
has attracted much attention in representation learning. The widely adopted contrastive learning optimizes the model by instance discrimination~\cite{simclr,robinson2020contrastive}. 
Specifically, it learns to embed the features of differently augmented versions of the same image to be similar, while being dissimilar if they came from different images. 
Some recent works~\cite{kim2020mixco,lee2020mix} have incorporated the idea of MixUp with contrastive learning. Our method is different from these methods in: ~\textit{1) They regard the mixed samples as queries and the original samples as keys, while we additionally consider a reverse operation to exchange their roles.  
2) In our method, the negative samples come from the same video as the positive samples, they can be viewed as hard negative samples, which is important in contarstive learning~\cite{robinson2020contrastive}.}

\noindent\textbf{Weakly-supervised temporal action localization}
aims to tackle TAL in the weakly-supervised setting.
UntrimmedNet~\cite{wang2017untrimmednets} is the pioneering work for it. 
In addition, there are some attempts~\cite{ma2020sfnet} to explore WTAL with only point-level action supervision, where each action instance is annotated with only a frame. 
Recently,~\cite{yang2021background} proposes a new WTAL setting with point-level background supervision, which annotates a frame in each background segment. 
In this work, we consider the former two types of supervision with more followers. 

Despite of different supervisions, most methods follow a localization-by-classification procedure, which formulates WTAL as a video classification task. Under this pipeline, an important  component is to select snippets with high probabilities of actions. In general, there are two groups of strategies: multiple instance learning (MIL)-based methods~\cite{CoLA,moniruzzaman2020action} and attention-based methods~\cite{STPN, Islam2021HAM-Net}. The former obtains the video-level scores from T-CAS by applying a pooling on the top-$k$ values for each class. The latter introduces the attention weights to eliminate background snippets. 
Recently, some WTAL methods have also noticed contrastive learning~\cite{CoLA,liexploring}. 
The difference between the above methods and ours is indeed obvious. ~\textit{They rely on pseudo-labels for defining positive and negative pairs. In contrast, we formulate it as an instance discrimination task, which is simpler and more generic.}


\vspace{-6mm}
\begin{figure}[t]
\centering
\includegraphics[width=1\linewidth]{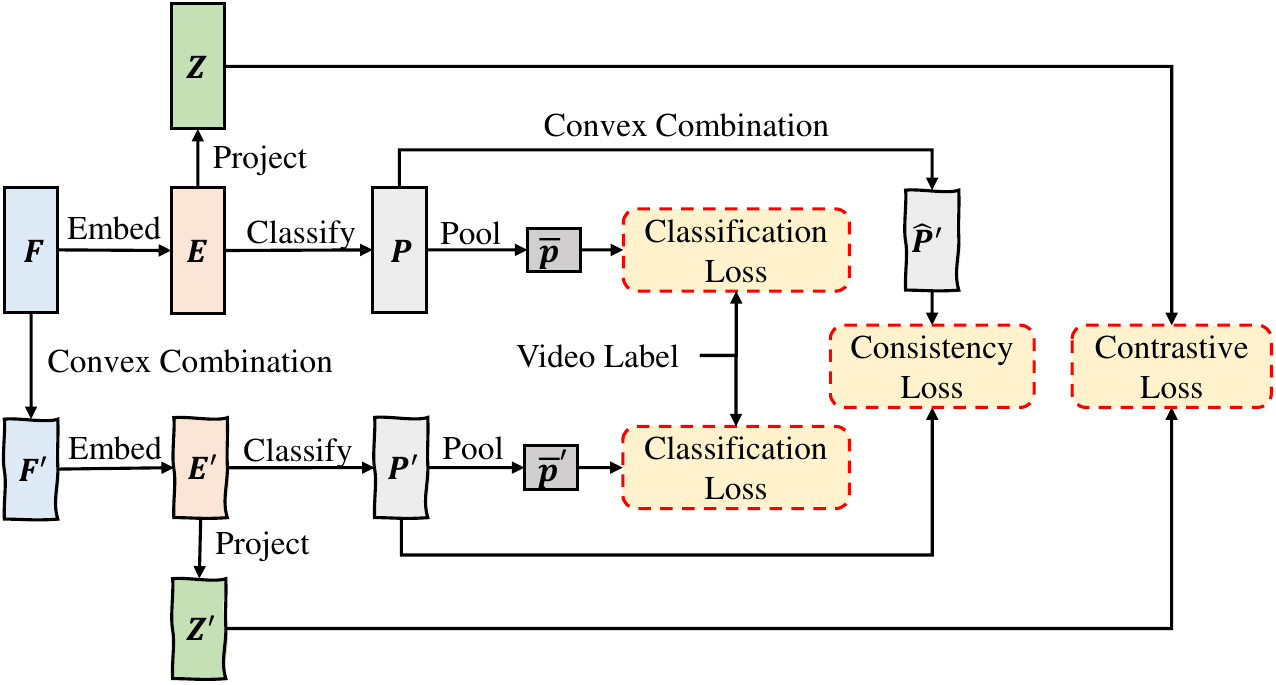}
\centering
\vspace{-6mm}
\caption{\textbf{Overview of  our method}. We first perform convex combination on input snippet sequence $\boldsymbol F$ to produce the augmented one $\boldsymbol F'$. The procedure is denoted as micro data augmentation. Then we simultaneously feed the $\boldsymbol F$ and $\boldsymbol F'$ into the model and compute four regularization loss terms. We call it macro-micro consistency regularization. }
\vspace{-3mm}
\label{fig:arch}
\end{figure}

\section{Our Method}
\vspace{-2mm}
In this section, we first review the basic pipeline of the mainstream WTAL methods which we adopt as our baselines. Then we elaborate on our proposed C$^3$BN, which contains two core components:  micro data augmentation and macro-micro consistency regularization, as depicted in Fig.~\ref{fig:arch}.

\vspace{-2mm}
\subsection{Preliminaries}\label{sec:recap}
\vspace{-1mm}
For an untrimmed video $\boldsymbol{V}$, we can only access its video-level label $\boldsymbol{y}=\{y_c\}_{c=1}^C$, where $C$ represents the number of action classes. A common practice is to employ a video classification model to predict the video-level label $\boldsymbol{y}$.  Specifically, we first divide $\boldsymbol{V}$ into $T$ non-overlapping snippets and then extract snippet-wise features $\boldsymbol F \in \mathrm{R}^{T\times D_f}$ using the pre-trained feature extractor.  
Since the extractor is not trained from scratch for the WTAL task, we further use several temporal convolution layers for mapping $\boldsymbol F$ to task-specific feature embedding $\boldsymbol E = [\boldsymbol e_1, ..., \boldsymbol e_T]\in \mathrm{R}^{T \times D_e}$. Afterward, $\boldsymbol E$ is fed into a snippet classifier to output a snippet prediction sequence  $\boldsymbol P = [\boldsymbol p_1, ..., \boldsymbol p_T] \in \mathrm{R}^{T\times C}$, where each snippet has its own scores $\boldsymbol p_t \in \mathrm{R}^{C}$. After this, we aggregate the snippet scores to obtain video-level scores $\boldsymbol {\bar p}  \in \mathrm{R}^{C} $. There are two main mechanisms in the literature for this purpose: MIL-based~\cite{CoLA} and attention-based~\cite{huang2021foreground}. The former applies a temporal top-k pooling to select high-score snippets  while the latter uses snippet-wise attention weights to aggregate snippets via attentive pooling. We refer to \textbf{Supplementary} for details.
Then, we formulate a video classification loss as follows:
\begin{equation} \small
\begin{aligned} 
\mathcal{L}_{cls} = -\frac{1}{C} \sum_{c=1}^{C} y_c \operatorname{log} \bar{p}_c.
\label{eq:video_loss}           
\end{aligned}
\end{equation}

\vspace{-3mm}
\subsection{Micro Data Augmentation}
\vspace{-1mm}
Upon the $T$ snippets $\{\boldsymbol f_t \}_{t=1}^{T}$, we perform convex combination on adjacent snippets to generate $T-1$ augmented snippets.
Formally, 
\begin{equation} \small
\begin{aligned} 
\boldsymbol f'_t = \alpha_t \boldsymbol f_t +  (1 - \alpha_t) \boldsymbol f_{t+1} \quad \forall t \in \{1, .., T-1\},
\label{eq:interp}        
\end{aligned}
\end{equation}
where $\boldsymbol f'_{t}$ is the child snippet and $\{\boldsymbol f_t, \boldsymbol f_{t+1 }\}$ are its parent snippets. The weight $\alpha_t \in (0, 1) $ is randomly sampled from a beta distribution $\mathrm{Beta}(\gamma, \gamma)$. 
Here $\gamma$ is a preset scalar.  
According to the temporal continuity of videos, the location of $\boldsymbol f'_t$ lies in between $t$ and $t+1$. Consequently, $\boldsymbol f'_t$ is always in front of $\boldsymbol f'_{t+1}$. Based on this principle, we stack the child snippets $\{ \boldsymbol f'_t \}_{t=1}^{T-1}$ along temporal dimension to form a 1D feature map dubbed $\boldsymbol F' \in \mathrm{R}^{(T-1)\times D_f}$.

\paragraph{Remark.} Notably, our proposed micro data augmentation is fundamentally different from the native MixUp. Specifically, MixUp randomly selects two images for mixing, which is not applicable to the snippets in WTAL as it will render the temporal locations of the mixed snippets undefined. In contrast, we intend to improve the boundary localization in WTAL and thus propose to mix the adjacent snippets, meanwhile taking advantage of the natural continuity within videos to define the locations of the mixed snippets. Besides, we propose various consistency regularization terms to encode more task-specific knowledge, which is also distinct from MixUp.

\vspace{-2mm}
\subsection{Macro-Micro Consistency Regularization} \label{sec:mmcr}
\vspace{-1mm}

To effectively exploit the child snippets during training, we derive three combinatorial rules to collaboratively regularize the learning procedure.

\paragraph{Video semantic consistency.}
The semantic label of each child snippet is expected to be the combination of its parents' labels. Then we can deduce that the video-level label of child sequence 
 $\boldsymbol F'$ is consistent with that of parent sequence $\boldsymbol F$. Therefore, it is feasible to utilize the known video-level labels to regularize the child sequences. Specifically, we feed $\boldsymbol F'$ into network and get a video classification loss named $L'_{cls}$ in the same form as $L_{cls}$. 
Since $\boldsymbol F'$ is a locally deformed version of $\boldsymbol F$, the usage of  $L'_{cls}$ is supposed to help to improve the robustness of the model. 
However, such macro regularization ignores the relationship between individual child snippet and parent snippet. To this end, we propose to further regularize the network from micro perspectives.  

\paragraph{Snippet prediction consistency.}
Inspired by MixUp, we encourage the model to behave linearly in-between adjacent snippets, thereby enhancing the ability of the model to classify adjacent snippets.
Without ground-truth labels, we propose to take the predictions of the snippets as their ``soft-labels''. Thereafter, we introduce a consistency regularization term to enforce the  soft-labels/prediction of a child snippet to be consistent with the same convex combination of soft-labels/predictions of its parent snippets. 

By feeding  $\boldsymbol F$ and $\boldsymbol F'$ into the model, we can obtain two snippet prediction sequences, namely $\boldsymbol P$ and $\boldsymbol P'$. 
To bridge $\boldsymbol P$ and $\boldsymbol P'$, we apply convex combination on $\boldsymbol P$ to obtain the shifted version of $\boldsymbol P$ dubbed $\boldsymbol {\widehat{P}'}$. Formally, 
\begin{equation} \small
\begin{aligned} 
\boldsymbol{\hat p'}_t = \alpha_t \boldsymbol p_t +  (1 - \alpha_t) \boldsymbol p_{t+1} \quad \forall t \in \{1, .., T-1\}.
\label{eq:interp_pred}        
\end{aligned}
\end{equation}
Then we apply the MSE loss to enforce the consistency between $\boldsymbol P'$ and $\boldsymbol {\widehat{P}}'$. Formally, 
\begin{equation} \small
\begin{aligned} 
\mathcal{L}_{cons} = \frac{1}{T-1}\sum_{t=1}^{T-1}||\boldsymbol p'_t - \boldsymbol{\hat{p}}'_t||_2^2.
\label{eq:consistency_loss}        
\end{aligned}
\end{equation}

\paragraph{Snippet feature contrastive-consistency.} 
Recent methods~\cite{xu2022multi,do2021clustering} demonstrate that feature contrastive learning is complementary to classifier learning. In light of these works, we propose to further regularize the intermediate features of the model via contrastive learning.
In particular, we develop a contrastive-consistency regularization that integrates the consistency regularization into the contrastive learning scheme.
By ``contrastive'', we mean that the model is forced to distinguish the parent/child snippets of each child/parent snippet from other parent/child snippets.
By ``consistency'', we mean that we enforce the model to learn the degree of proximity between child and parent snippets. Here we jointly achieve both of goals, and let the snippet features be aware of the relative similarity in-between adjacent snippets in comparing with other snippets. Consequently, the features would gradually capture necessary fine-grained discriminability to distinguish subtle differences between adjacent snippets. 

First, to avoid the conflict between the instance-based contrastive learning and the underlying semantics within the feature embedding $\boldsymbol E$, we append a projection head~\cite{simclr}, comprised by a $\operatorname{FC}$ layer and a $\operatorname{L2}$ normalization, to map $\boldsymbol E$ into a low-dimensional unit hypersphere in which the contrastive learning is performed. As a result,  $\boldsymbol E$ serves as a medium for information transfer between the classifier and the projection head, allowing the classifier to leverage disciminative fine-grained patterns captured by the projection head to extract accurate class-level patterns. 
Let us denote the output by $\boldsymbol Z = [\boldsymbol z_1, .., \boldsymbol z_t] \in \mathrm{R}^{T \times D_z}$. Here $D_z$ is the number of channels with $D_z < D_f$.
Similarly, we can obtain the counterpart of $\boldsymbol E'$ denoted by $\boldsymbol Z' = [\boldsymbol z'_1, .., \boldsymbol z'_t]$. 

Then, taking each child snippet $\boldsymbol z'_t $ as a query, we define that: 1) its parent snippets $\boldsymbol z_t$ and $\boldsymbol z_{t+1}$ are its semi-positive keys with the probability of $\alpha_t$ and $1 - \alpha_t$, respectively. 2) other snippets in $\boldsymbol Z$ are negative keys. Thus we can construct a soft contrastive loss as follows:
\begin{equation} \small
\begin{aligned} 
\mathcal{L}_{cont} =- \frac{1}{T-1}\sum_{t=1}^{T-1}  \alpha_t  \operatorname{log} \frac{\exp({\boldsymbol z'}_t^\top \boldsymbol z_t/\rho)  
}{\sum_{\tau=1}^T\exp({\boldsymbol z'}_t^\top \boldsymbol z_\tau/\rho)} + \\  (1 - \alpha_t) \operatorname{log} \frac{\exp({\boldsymbol z'}_t^\top \boldsymbol z_{t+1}/\rho)  
}{\sum_{\tau=1}^T\exp({\boldsymbol z'}_t^\top \boldsymbol z_\tau/\rho)} ,
\label{eq:contrastive}        
\end{aligned}
\end{equation}
where $\rho$ is the temperature coefficient. 

Eq.\eqref{eq:contrastive} only considers the unilateral reference from $\boldsymbol Z'$ to $\boldsymbol Z$. To explore more fine-grained patterns and enhance the consistency regularization between $\boldsymbol Z$ and $\boldsymbol Z'$, we propose a bilateral reference mechanism to further take the reference from $\boldsymbol Z$ to $\boldsymbol Z'$ into consideration. That is, we treat the elements of $\boldsymbol Z $ as queries and the elements of  $\boldsymbol Z' $ as keys. Meanwhile, the snippet-to-snippet relations remain unchanged. As a result, we can compute another contrastive loss dubbed $\mathcal{L}'_{cont}$ in a similar way to Eq.\eqref{eq:contrastive}.  

\paragraph{Remark.} 
The three regularization terms are introduced to work collaboratively, comprehensively promoting the model training from~\textit{macro view} to~\textit{micro view}  and from~\textit{low-level representations} to~\textit{high-level semantics}. 
Intuitively, these constraints together will encourage the model to exploit the various relationships between parent and child snippets, eventually facilitating the exploration of fine-grained distinctions between adjacent snippets.  We will show later in Sec.~\ref{sec:exp} the efficacy and compatibility of these terms. 
It is noteworthy that the concrete differences between our proposed regularization strategies and related methods are highlighted in Sec.~\ref{sec:related_work}.

\vspace{-2mm}
\section{Experiments}
\label{sec:exp}
In this section, we empirically validate the effectiveness of  C$^3$BN. 
Due to space limitation, we refer to \textbf{Supplementary} for more details for all experiments.  Moreover, we only report the results on WTAL with video-level supervision and refer to \textbf{Supplementary} for WTAL with point-level supervision.

\vspace{-2mm}
\subsection{Dataset and Metrics}
\vspace{-1mm}
\textit{THUMOS14}~\cite{THUMOS14}  contains untrimmed videos with 20 classes.
By convention, we use the 200 videos in validation set for training and 213 videos in test set for evaluation. \textit{ActivityNet v1.3}~\cite{caba2015activitynet} is a large-scale dataset with 200 categories. By convention, we train on the training set with 10, 024 videos and test on validation set with 4, 926 videos. The mean Average Precision (mAP) values under different temporal intersection
over union (tIoU) thresholds are used as metrics.

\vspace{-2mm}
\subsection{Ablation study}
\vspace{-1mm}

\paragraph{Effectiveness on different baselines}
To validate the  generic effectiveness of C$^3$BN, we incorporate C$^3$BN into different WTAL methods. Specifically, we plug C$^3$BN into four baselines, including the aforementioned MIL-based baseline (named \textbf{MIL}) and three off-the-shelf well-performing approaches, \ie,  \textbf{BaSNet}~\cite{lee2020background}, \textbf{FACNet}~\cite{huang2021foreground}, and recent \textbf{DELU}~\cite{chen2022dual}(ECCV 2022). -
Table~\ref{table:THUMOS14} shows the performance comparison. We can observe that C$^3$BN consistently improves the performance of all baselines by 7.4\%, 3.9\%,  1.5\%, and 1.1\% on AVG mAP for MIL, BaSNet, FACNet, and DELU, respectively. 
These results clearly confirm the generalization ability of C$^3$BN.

\begin{table}[!t]
	\begin{center}
        \setlength{\tabcolsep}{7pt}
        \vspace{-3mm}
	\caption{Comparisons of performance on THUMOS14. The AVG represents average mAP under IoU thresholds of 0.1:0.7. We re-implement all the adopted baselines for C$^3$BN.  }
 \vspace{-2mm}
		\label{table:THUMOS14}
		\resizebox{\linewidth}{!}{
			\begin{tabular}{l|ccccccc|l}
				\hline
				\multirow{2}{*}{Method}  & \multicolumn{7}{c|}{mAP @ IoU (\%)}  & \multirow{2}{*}{AVG} \\
				\cline{2-8}
				& 0.1 & 0.2 & 0.3 & 0.4 & 0.5 & 0.6 & 0.7 & \\
				\hline
				WUM  \cite{lee2021weakly} & 67.5 & 61.2 & 52.3 & 43.4 & 33.7 & 22.9 & 12.1 & 41.9 \\
				AUMN \cite{luo2021action} & 66.2 & 61.9 & 54.9 & 44.4 & 33.3 & 20.5 & 9.0 & 41.5 \\
				CoLA \cite{CoLA} & 66.2 & 59.5 & 51.5 & 41.9 & 32.2 & 22.0 & 13.1 & 40.9 \\
				UGCT \cite{yang2021uncertainty} & 69.2 & 62.9 & 55.5 & 46.5 & 35.9 & 23.8 & 54.0 & 43.6 \\
				DCC \cite{liexploring} & 69.0 & 63.8 & 55.9 & 45.9 & 35.7 & 24.3 & 13.7 & 44.0 \\
				RSKP \cite{huang2022weakly} & 71.3 & 65.3 & 55.8 & 47.5 & 38.2 & 25.4 & 12.5 & 45.1 \\
				ASM-Loc \cite{he2022asm} & 71.2 & 65.5 & 57.1 & 46.8 & 36.6 & 25.2 & 13.4 & 45.1 \\
                Li~\etal~\cite{li2022forcing} & 69.7&  64.5& 58.1& \textbf{49.9} &  39.6  & 27.3 & 14.2 & 46.1 \\
				\cline{1-9}
				MIL  & 56.0 & 46.4 & 37.3 & 30.3 & 22.0 & 15.0 & 8.2 & 30.7 \\
				\rowcolor{gray!10}  + C$^3$BN  & 63.0 & 56.7 & 48.0 & 39.8 & 29.9 & 19.2 & 10.2 & 38.1$_{\textcolor{blue}{+7.4}}$ \\
				\cline{1-9}
				BaSNet~\cite{lee2020background}  & 62.0 & 54.6 & 44.6 & 35.7 & 25.9 & 17.0 & 8.9 & 35.5 \\
				\rowcolor{gray!10} + C$^3$BN  & 64.3 & 58.4 & 49.7 & 40.6 & 30.8 & 19.9 & 12.1  & 39.4$_{\textcolor{blue}{+3.9}}$ \\
				\cline{1-9}
				FACNet~\cite{huang2021foreground}  & 71.8 & 64.0 & 53.7 & 42.5 & 30.7 & 20.9 & 12.2   & 42.3 \\
				\rowcolor{gray!10} + C$^3$BN  & \textbf{72.6} & \textbf{66.5} & 56.4 & 43.8 & 32.6 & 21.0 & 12.7& 43.7$_{\textcolor{blue}{+1.5}}$ \\
				\cline{1-9}
    			DELU~\cite{chen2022dual}  & 70.1 & 64.5 & 56.0 & 47.6 & 40.2 & 27.8 & 15.0 & 45.9 \\
				\rowcolor{gray!10} + C$^3$BN  & 71.6 & 66.0 & \textbf{58.2} & 49.3 & \textbf{41.0} & \textbf{27.9} & \textbf{15.3} & \textbf{47.0}$_{\textcolor{blue}{+1.1}}$ \\
				\hline
		\end{tabular}}
    \vspace{-3mm}
	\end{center}
\end{table}

\begin{table}[!t]
\setlength{\tabcolsep}{17pt}
\vspace{-5mm}
\caption{Results on ActivityNet v1.3. AVG indicates the average mAP at IoU thresholds 0.5:0.05:0.95.}
\vspace{-2mm}
		\begin{center}
		\resizebox{0.99\linewidth}{!}{
		\label{table:activity1.3}
				\begin{tabular}{l|cccl}
					\hline
					\multirow{2}{*}{Method} & \multicolumn{4}{c}{mAP @ IoU} \\
					\cline{2-5}
					& 0.5 & 0.75 & 0.95 & AVG \\
					\hline
					WUM \cite{lee2021weakly} & 37.0 & 23.9 & 5.7 & 23.7\\
     				AUMN  \cite{luo2021action} & 38.3 & 23.5 & 5.2 & 23.5 \\
					UGCT \cite{yang2021uncertainty} & 39.1 & 22.4 & 5.8 & 23.8 \\
					DCC \cite{liexploring} & 38.8 & 24.2 & 5.7 & 24.3 \\
					RSKP \cite{huang2022weakly} & 40.6 & 24.6 & 5.9 & 25.0 \\
					ASM-Loc \cite{he2022asm} & 41.0 & 24.9 & 6.2 & 25.1 \\
					\hline
					BaSNet  &35.6 &21.0 &5.3 &21.7 \\
					+ C$^3$BN  &37.3 &22.4 &5.4 &23.0$_{\textcolor{blue}{+1.3}}$ \\
					\hline
					FACNet  \cite{huang2021foreground}  &40.1 &24.2 &5.8 &24.7 \\
					\rowcolor{gray!10} + C$^3$BN &\bf{45.2} &\bf{26.9} &\bf{5.9} &\bf{27.3$_{\textcolor{blue}{+2.6}}$} \\
            \hline
			\end{tabular}}
   \vspace{-3mm}
		\end{center}
\end{table}

\begin{table}[!t]
\setlength{\tabcolsep}{12pt}
\vspace{-4mm}
\caption{Ablation studies of the proposed regularization terms on THUMOS14. }
\vspace{-2mm}
\label{table:ablation_thu14}
\begin{center}
\resizebox{0.99\linewidth}{!}{
\begin{tabular}{c|c|c|cc|cc}
\hline
\multicolumn{1}{c|}{\multirow{2}{*}{\#}} &
\multicolumn{4}{c|}{Loss terms} & 
\multicolumn{2}{c}{Baselines} \\
\cline{2-7}
 &$\mathcal{L}^{'}_{cls}$  &$\mathcal{L}_{cons}$   &$\mathcal{L}_{cont}$  &$\mathcal{L}'_{cont}$  & BaSNet & FACNet \\
\hline 
1 & &  &  &  & 35.5  & 42.3 \\
2 &$\surd$ &  &  &  & 35.9  & 42.6 \\
3&  &$\surd$&  &  & 37.7 & 43.0 \\
4 &  &  & $\surd$  & $\surd$  & 36.5  & 42.8\\
\hline 
5 & $\surd$ &  $\surd$ &  &   & 38.0 & 43.0 \\
6 & $\surd$ &  &  $\surd$  &   $\surd$  & 36.8 & 42.9  \\
7 &  & $\surd$ &  $\surd$  &   $\surd$  & 38.8 & 43.5 \\
\hline 
8 &  $\surd$  & $\surd$ &   &   $\surd$  & 39.1 & 43.4 \\
9 &  $\surd$ & $\surd$ &  $\surd$ &     & 38.9 & 43.5 \\
\hline
\rowcolor{gray!10}  10 &  $\surd$ & $\surd$ &  $\surd$ &  $\surd$   & 39.4 & 43.7  \\
\hline
\end{tabular}
}
\vspace{-3mm}
\end{center}
\end{table}

\begin{table}[!t]
\vspace{-5mm}
\caption{Contribution of C$^3$BN to {\ding{182}} boundary localization and {\ding{183}} proposal evaluation. ``{\ding{182}}Base+{\ding{183}}Base'' represents the baseline, \ie, BaSNet and FACNet. ``{\ding{182}}C$^3$BN+{\ding{183}}Base'' indicates that we combine our {\ding{182}} and the {\ding{183}} of baseline. Likewise for ``{\ding{182}}Base+{\ding{183}}C$^3$BN'' and ``{\ding{182}}C$^3$BN+{\ding{183}}C$^3$BN''. }
\vspace{-2mm}
\label{table:boundary}
\centering
\begin{center}
\resizebox{1\linewidth}{!}{
\begin{tabular}{c|c|c|c|c}
		\hline
	    & {\ding{182}}Base+{\ding{183}}Base & {\ding{182}}C$^3$BN+{\ding{183}}Base &  {\ding{182}}Base+{\ding{183}}C$^3$BN & {\ding{182}}C$^3$BN+{\ding{183}}C$^3$BN  \\
		\hline
		 BaSNet &35.5& 38.0 & 37.1 &39.4  \\
		 FACNet &42.3& 43.2 & 42.8 &43.7  \\
        \hline
\end{tabular}
}
\vspace{-5mm}
\end{center}
\end{table}

\paragraph{Contribution of each regularization term}  
Our C$^3$BN introduces several regularization/loss terms during training. 
To verify the contribution of each regularization term, we conduct a detailed analysis in Table~\ref{table:ablation_thu14}. 
Here, we regard BaSNet and FACNet as the baselines to conduct the ablation study due to their favorable efficiency and flexibility.
Comparing the rows \#1-4, we can see that each regularization term contributes to the performance. Furthermore, it can be seen that the micro consistency regularization terms (\ie, $\mathcal{L}_{cons}$, $\mathcal{L}_{cont}$ and $\mathcal{L}'_{cont}$ ) bring larger gains than the macro term (\ie, $\mathcal{L}'_{cls}$). This indicates that fine-grained information is pretty important in WTAL. 

\paragraph{Complementarity of regularization terms}  
In rows \#5-6 of Table~\ref{table:ablation_thu14}, we evaluate the performance of combining any two of the regularization terms. We can see that combining two terms consistently outperforms each of them. Moreover, after combining all the terms, the model obtains the best performance, as shown in row \#10. These results evidently demonstrate the complementary relations of the regularization terms.      

\paragraph{Effectiveness of bilateral reference mechanism} We propose a bilateral reference mechanism in the snippet feature contrastive-consistency regularization, resulting in two loss terms (\ie, $\mathcal{L}_{cont}$ and $\mathcal{L}'_{cont}$). In rows \#8-9 of Table~\ref{table:ablation_thu14}, we provide the results where only one of them is used. It can be seen that the combination of $\mathcal{L}_{cont}$ and $\mathcal{L}'_{cont}$ outperforms using only of them, validating the superiority of our proposed bilateral reference mechanism. 

\paragraph{Necessity of projection head}
We adopt a projection head to transform the embeddings into a new latent space so that the instance-based contrastive learning would not directly hurt the inherent semantics of the embeddings. To show the necessity of our design, we conduct an experiment where the projection head is removed. The experimental results show that it leads to a performance degradation of 1.1\% on BaSNet (from 39.4\% to 38.3\%) and 0.9\% on FACNet (from 43.7\% to 42.8\%). This evidently justifies that the projection head is essential in our method.   

\vspace{-2mm}
\subsection{Comparisons with state-of-the-arts (SOTAs)}
\vspace{-1mm}
Table~\ref{table:THUMOS14} and Table~\ref{table:activity1.3} show the comparison between our method and previous approaches on THUMOS14 and ActivityNet v1.3, respectively. It can be seen that after integrating recent strong WTAL baselines, our method achieves the SOTA performances on both datasets. In \textbf{Supplementary}, we also report the results on~\textit{ActivityNet v1.2}~\cite{caba2015activitynet} (a subset of ActivityNet v1.3), which is used in some previous methods~\cite{chen2022dual,li2022forcing}.

\vspace{-2mm}
\subsection{Evaluation for Motivation}
\vspace{-1mm}
In this section, we provide experimental results to deliver more insights for our motivation (depicted in Fig.~\ref{fig:issue}).

To begin with, we investigate the contribution of C$^3$BN to  ``{\ding{182}} boundary localization''  and ``{\ding{183}} proposal evaluation'' respectively in Table~\ref{table:boundary}. To be specific, in the test phase, we alternately replace the results of {\ding{182}} and {\ding{183}} of baseline with that of baseline+C$^3$BN (ours).
As we can see, the performances significantly increase (from 35.5\% to 38.0\% on BaSNet and from 42.3\% to 43.2\% on FACNet ) once replacing the {\ding{182}} of baselines with our {\ding{182}}.  As a comparison, the replacement on {\ding{183}} only brings the performance to 37.1\% on BaSNet and  42.8\% on FACNet. These results indicate that C$^3$BN is particularly beneficial for boosting boundary localization. 

To understand how C$^3$BN improves the boundary localization, we first compute the absolute score difference between each pair of adjacent snippets, \ie, $\boldsymbol d_t=|\boldsymbol p_{t+1}-\boldsymbol p_t|$. Next, we calculate the average entropy of $\boldsymbol d_t$ of all pairs, \ie, $H(\boldsymbol d_t) = \text{mean}(-\boldsymbol d_t \log \boldsymbol d_t)$. Intuitively, the $H(\boldsymbol d_t)$ reflects the distribution of the differences, \ie, the smaller the $H(\boldsymbol d_t)$, the more polarized the differences, and the more discriminative and confident the model is about the relations of adjacent snippets. Experimental results show that the corresponding $H(\boldsymbol d_t)$ of BaSNet and FACNet is 0.0876 and 0.0673 respectively while that  of ``BaSNet + C$^3$BN'' and ``FACNet + C$^3$BN''  is 0.0684 and 0.0592 respectively. 
Hence, we conjecture that the boundary localization is improved because C$^3$BN renders the model more confident (discriminative) in distinguishing the relations of adjacent snippets. 

In \textbf{Supplementary}, we provide extensive qualitative results to demonstrate it.   

\vspace{-2mm}
\section{Conclusion}
\vspace{-2mm}
In this paper, we propose a universal training strategy dubbed C$^3$BN for weakly-supervised action localization. Concretely, C$^3$BN first produces new snippets by convex combination between adjacent snippets, and then uses them to regularize the model with three regularization terms, \ie, video semantic consistency, snippet prediction consistency and snippet feature contrastive-consistency. The empirical results validate that C$^3$BN is applicable to various WTAL methods with video-level supervision and point-level supervision, and helps establish the new SOTA results on all the evaluated datasets.

{\footnotesize
\bibliographystyle{IEEEbib}
\bibliography{icme2023template}
}

\clearpage
\appendix
{
  \hypersetup{linkcolor=black}
  \tableofcontents
}

\section{Additional Details of Our Method} \label{sec:baseline}

\paragraph{Additional details of baseline.}
Given the video labels, we first aggregate the snippet scores to obtain video class scores for computing a video classification loss. There are two main strategies in the literature for this purpose: MIL-based methods~\cite{wang2017untrimmednets,CoLA} and attention-based methods~\cite{background_modeling,liu2019completeness}.

The  MIL-based methods average the top-$k$ snippet \textit{logit} scores (dubbed $\boldsymbol S = [\boldsymbol s_1, ..., \boldsymbol s_T] \in \mathrm{R}^{T\times C}$)  along temporal dimension for each class to build the video class score:
\begin{equation}
\begin{aligned} 
 {\bar s_{c} =\frac{1}{k}\max_{\begin{subarray}{}\boldsymbol{l}\subset \{1,..,T\} \\\quad |\boldsymbol{l}|=k \end{subarray} } \sum_{\tau \in \boldsymbol{l}}{s_{\tau,c}}} \quad \forall  c\in \{1,.., C\},
\label{eq:weight_sum}        
\end{aligned}
\end{equation}
where $k$ is a hyper-parameter proportional to the video length $T$, \textit{i.e.}, $k = \max(1, T//r)$, and $r$ is a pre-defined parameter. 
Thereafter, we obtain the probability for each class by applying the $\operatorname{Softmax}$ function to the aggregated scores:
\begin{equation}
\begin{aligned} 
\bar p_c = \frac{\exp(\bar s_c)}{\sum_{i=1}^{C}{\exp(\bar s_i)}} \quad \forall  c\in \{1,.., C\}.
\end{aligned}
\end{equation}

The attention-based methods first learn a set of snippet-wise attention weights (dubbed $\boldsymbol \lambda = [\lambda_1, ..., \lambda_T] \in \mathrm{R}^{T}$). Then the attention weights are used to aggregate snippet-level scores into video-level scores as follows,  

\begin{equation}
\begin{aligned} 
\boldsymbol{\bar p} = \frac{1}{\sum_{t=1}^{T} \lambda_t} \sum_{t=1}^{T} \lambda_t  \boldsymbol{ p_t} .
\end{aligned}
\end{equation}

\begin{figure}[t]
\centering
\includegraphics[width=0.9\linewidth]{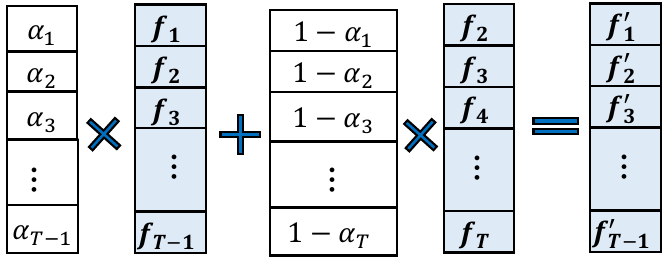}
\centering\caption{\textbf{Illustration of micro data augmentation.} The $\{ \boldsymbol f_t \}_{t=1}^{T-1}$ indicates the parent  sequence and the $\{ \boldsymbol f'_t \}_{t=1}^{T-1}$ represents the child sequence.}
\label{fig:CC}
\end{figure}
\begin{figure}[t]
\centering
\includegraphics[width=0.9\linewidth]{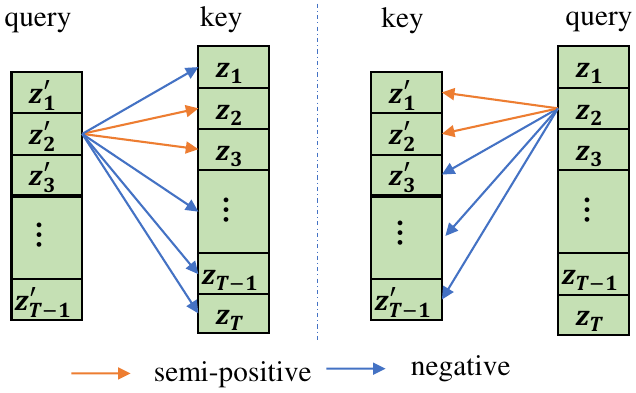}
\centering
\caption{\textbf{Illustration of snippet feature contrastive-consistency.} It contains two parts: the child snippets are query while the parent ones are key (left);  the parent snippets are query while the child ones are key (right). }
\label{fig:CL}
\end{figure}

\paragraph{Illustration of micro data augmentation and snippet feature contrastive-consistency . }  In Fig.~\ref{fig:CC} and  Fig.~\ref{fig:CL}, we shows more details for micro data augmentation and snippet feature contrastive-consistency respectively.

\paragraph{Training objective.}
To train the entire model in an end-to-end fashion, we optimize the following loss 
\begin{equation} \small
\begin{aligned} 
\mathcal{L} =  \mathcal{L}_{base} + \lambda_1 \mathcal{L}'_{cls} + \lambda_2 \mathcal{L}_{cons} + \lambda_3  (\mathcal{L}_{cont} + \mathcal{L}'_{cont}).
\label{eq:loss_sum}        
\end{aligned}
\end{equation}
Here $\lambda_*$ indicates the weight term. 
$ \mathcal{L}_{base}= \mathcal{L}_{cls} + \mathcal{L}_{other}$ represents the objective function of the baseline, where $\mathcal{L}_{other}$ represents the sum of other losses apart from $\mathcal{L}_{cls}$.

\section{Experiments}
\label{sec:exp}
\subsection{Implementation Details}
\paragraph{Implementation details of baselines}
\textbf{MIL} is a simple MIL-based baseline, which has been introduced Sec.~\ref{sec:baseline}. Briefly, it first performs snippet classification over the input sequences to obtain T-CAS, then utilizes a top-k pooling operation following~\cite{UntrimmedNet} to build video-level scores. At last, it minimizes a cross-entropy loss with the video-level labels. We implement MIL based on the code provided by~\cite{CoLA}, except that we disable the contrastive loss used in~\cite{CoLA}.    
\textbf{BaSNet}~\cite{lee2020background} introduces an additional class to the snippet-level classifier for modeling background. Besides, it utilizes a class-agnostic attention layer to highlight foreground snippets and suppress the background snippets.
The performance for BaSNet implemented by~\cite{lee2020background} is not stable on THUMOS14. Thus, we re-implement it on the basis of the code of~\cite{ma2021weakly,huang2021foreground}. We refer to our released code for details.  \textbf{FACNet}~\cite{huang2021foreground} proposes to force the foreground score output by the snippet-level classifier and  that output by the attention layer to be consistent. We implement the FACNet in a similar way to the BaSNet. \textbf{DELU}~\cite{chen2022dual} extends the traditional
paradigm of evidential deep learning to adapt to the weakly-supervised multi-label classification goal. \textbf{SF-Net}~\cite{ma2020sfnet} mines pseudo action and background frames by adaptively expanding each annotated single frame to its nearby frames. \textbf{LACP}~\cite{lee2021LACP} takes the points as seeds and searchs for the optimal sequence that is likely to contain complete action instances while agreeing with the seeds. For DELU, SF-Net and LACP, we use their official code to implement them. All the baselines are implemented on the Pytorch library.  

\paragraph{Training details}  For the feature extraction, we first sample RGB frames at 25 fps for each video and apply the TV-L1 algorithm~\cite{zach2007duality} to generate optical flow frames. Then, we divide each video into non-overlapping snippets with consecutive 16 frames. Thereafter, we perform the I3D network~\cite{I3D} pre-trained on the Kinetics dataset~\cite{kinetics_dataset} to obtain the snippet-level feature. The proposed C$^3$BN and the baseline models are jointly trained in an end-to-end manner. 
Here, we only provide the details about the specific hyperparameters of C$^3$BN. The temperature $\rho$ is set as $0.1$ and the output dimension of projection head $D_z$ is set as $128$ and the $\gamma$ is set as $2$. Since the amplitudes of basic loss in different baselines are different, the loss weights $\lambda_2$ and $\lambda_3$ are set differently on different baselines, except that the $\lambda_1$ is always set by $1$. On BaSNet, the $\lambda_2$ and $\lambda_3$ are set as $10$ and $0.1$, respectively. On FACNet, the $\lambda_2$ and $\lambda_3$ are set as $1$ and $0.2$, respectively. On DELU, the $\lambda_2$ and $\lambda_3$ are set as $0.1$ and $0.3$, respectively. On SF-Net, the $\lambda_2$ and $\lambda_3$ are set as $10$ and $0.2$, respectively. On LACP, the  $\lambda_2$ and $\lambda_3$ are set as $2$ and $0.1$, respectively. All experiments are conducted on one GTX 3090 GPU (24 GB). 

\paragraph{Inference details} The proposed C$^3$BN is a training strategy, which introduces no overhead in the test phase. As addressed in main paper, for existing WTAL methods, there exist two core procedures in the test phase, \ie, boundary localization and proposal evaluation. Despite of this, the actual test paradigms in different baselines are slightly different. We hereby take our implementation on BaSNet as an example for illustration.  In the inference stage, we first threshold on the video-level scores to determine the action categories. And then for the selected action class, we apply a threshold strategy on the T-CAS to obtain action proposals. After obtaining the action proposals, we calculate the class-specific score for each proposal using the outer-inner-contrastive technique~\cite{shou2018autoloc}. To enrich the proposal pool, multiple thresholds are applied. The Non-Maximum Suppression (NMS) is used to remove duplicated proposals, where SoftNMS~\cite{bodla2017soft} is particularly adopted.

\paragraph{Additional details of datasets.}
For W-TAL with video-level supervision, we conduct experiments on two popular benchmark datasets: \textit{THUMOS14}~\cite{THUMOS14} and \textit{ActivityNet v1.3}~\cite{caba2015activitynet}.   
THUMOS14 contains untrimmed videos with 20 classes. 
The video length varies from a few seconds to several minutes and multiple action instances may exist in a single video, which makes it very challenging. By convention, we use the 200 videos in validation set for training and the 213 videos in test set for evaluation. 
ActivityNet v1.3 is a large-scale dataset with $200$ categories. Since the annotations for the test set are not released, following the common practice, we train on the trainining set with $10,024$ videos and test on validation set with $4,926$ videos. \textit{ActivityNet v1.2}~\cite{caba2015activitynet} is a subset of ActivityNet v1.3, and covers 100 action categories with 4, 819 and 2, 383 videos in the training and validation sets, respectively 

For W-TAL with point-level supervision, three public datasets are commonly used, including \textit{THUMOS14}, \textit{BEOID}~\cite{calway2015discovering}, and \textit{GTEA}~\cite{lei2018temporal}.   
GTEA contains 28 videos of 7 fine-grained types of activities in the kitchen. There are 58 videos from 30 action classes in BEOID. We follow~\cite{ma2020sfnet} to split the training and test sets.

\begin{figure*}[!ht]
\centering
\includegraphics[width=\linewidth]{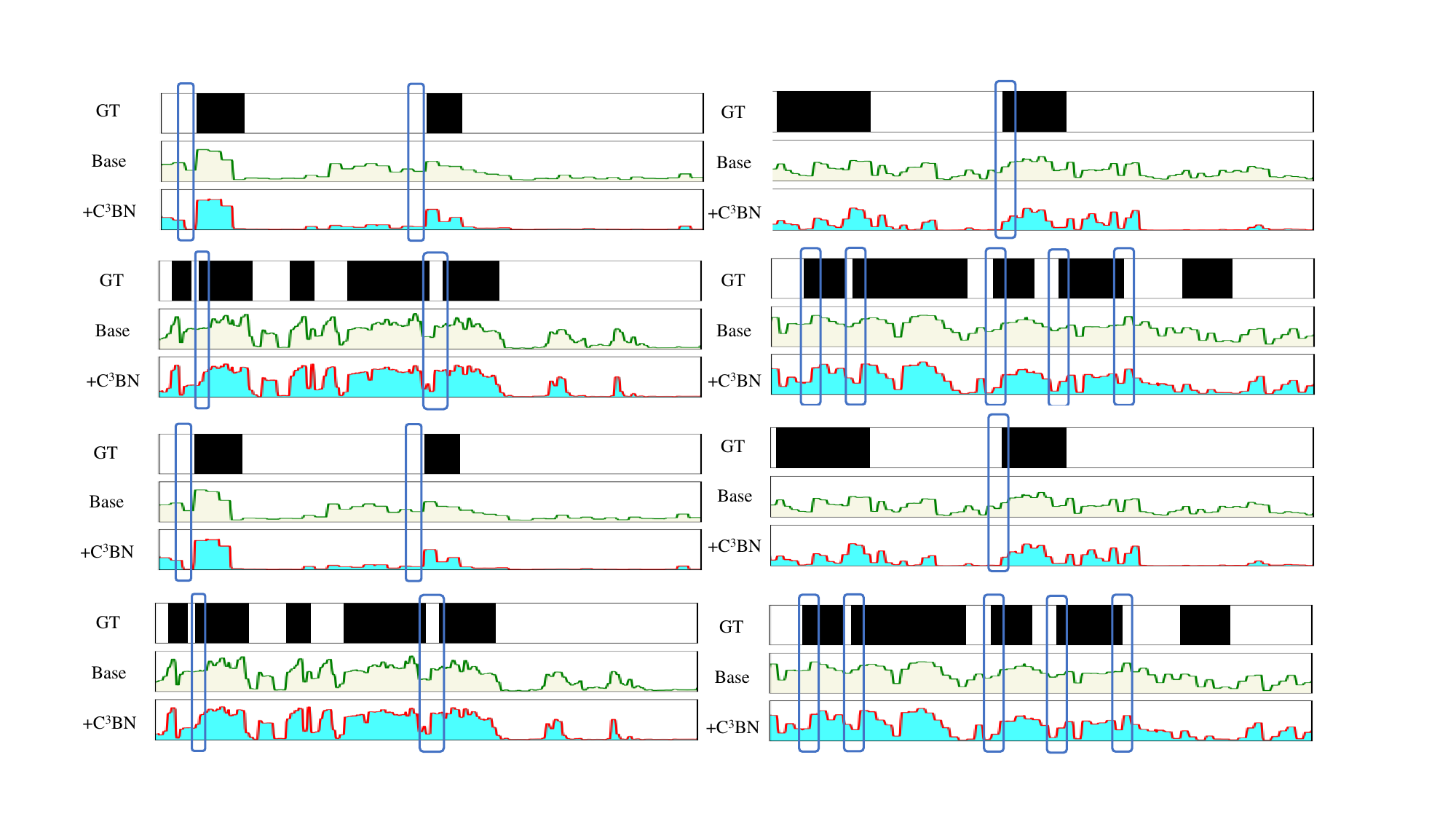}
\centering
\caption{\textbf{Qualitative results of T-CAS.} ``GT'' denotes ground-truth annotation. ``Base'' denotes the T-CAS predicted by BaSNet while ``+C$^3$BN'' denotes that predicted by ``BaSNet + C$^3$BN''. The solid boxes indicate some noteworthy regions. }
\label{fig:vis}
\end{figure*}
\begin{figure}[!ht]
\centering
\includegraphics[width=\linewidth]{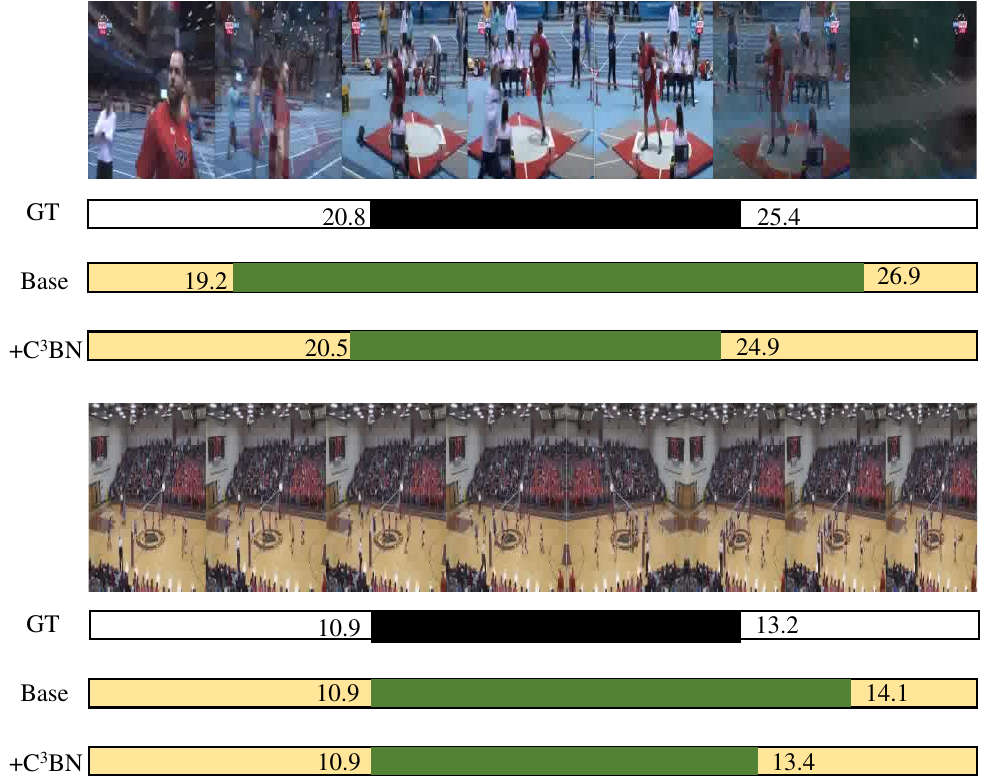}
\centering
\caption{\textbf{Qualitative results of proposals.} ``GT'' denotes ground-truth annotation. ``Base'' denotes the proposals output by BaS-Net while ``+C$^3$BN'' denotes the proposals generated by ``BaS-Net + C$^3$BN''. }
\label{fig:vis_prop}
\end{figure}

\begin{figure}[!ht]
\centering
\includegraphics[width=1\linewidth]{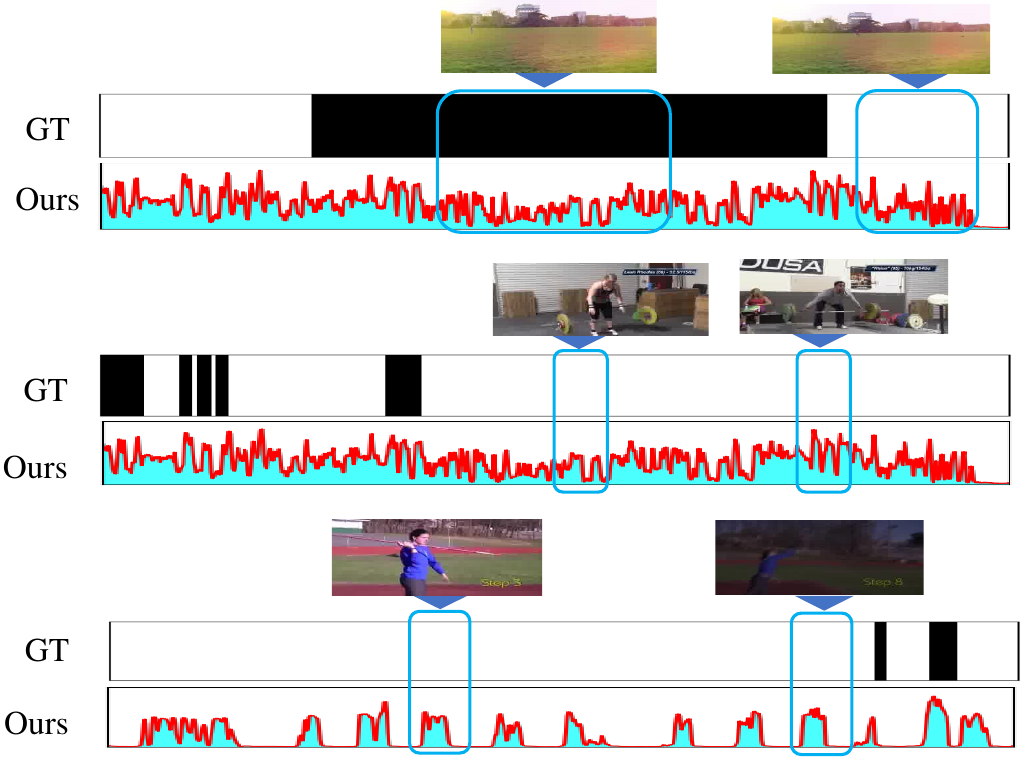}
\centering
\caption{\textbf{Samples of failure cases.}  We highlight the regions with wrong predictions by dashed boxes.}
\label{fig:vis_fail}
\end{figure}

\subsection{Qualitative Results} 

To gain further insights, we visualize a couple of samples for comparing the snippet-level predictions of the baseline model and that of our method in Fig.~\ref{fig:vis}.  From the solid boxes of Fig.~\ref{fig:vis}, it can be seen that our method is able to generate accurate action boundaries, while the baseline suffers from poor discrimination around boundaries. These visualized examples evidently and intuitively verify our motivation. 

In Fig.~\ref{fig:vis_prop}, we show some visualized results of the proposal generation. It can be seen that compared with the baseline model, the boundary of the proposals generated by our method is closer to the ground-truth action boundaries. This further demonstrates the superiority of our method in boundary localization.

Additionally, Fig.~\ref{fig:vis_fail} illustrates some failure cases of our method. The failure cases are caused by 1) low quality of images (see the top row);  2) ambiguous action boundary annotation (see the middle row) ; 3) indistinguishable body motions (see the bottom row). These challenging cases are our future work. 

 \begin{table}[ht]
\centering
\caption{Comparison on THUMOS14, GTEA and BEOID. AVG indicates the average mAP at IoU thresholds 0.1:0.7.}
\label{table:comparison_small_datasets}
\begin{center}
\resizebox{\linewidth}{!}{
\begin{tabular}{c|l|cccc|l}
		\hline
			Dataset & Method &0.1& 0.3 &0.5 &0.7 &AVG  \\
			\hline
			\multirow{5}{*}{THUMOS}
			&DC~\cite{ju2021DC}  & 72.8 &58.1 &34.5 &11.9 & 44.3   \\
			\cline{2-7}
            &SF-Net~\cite{ma2020sfnet} & 69.9 & 53.6 & 29.9 &10.0 & 40.9   \\
             &  Our C$^3$BN & 73.8 &57.3 &30.7 &10.3 & 43.3$_{\textcolor{blue}{+2.4}}$   \\
            \cline{2-7}
            &LACP~\cite{lee2021LACP} & 75.5 &64.0 &44.5 &20.6 & 52.3   \\
             &   Our C$^3$BN & \textbf{76.0} & \textbf{65.7} & \textbf{47.6} & \textbf{22.6} & \textbf{54.1}$_{\textcolor{blue}{{+1.8}}}$    \\
			\hline
			\multirow{3}{*}{GTEA}
			&DC~\cite{ju2021DC} &\bf{59.7} &38.3&21.9&18.1&33.7   \\
			\cline{2-7}
            &SF-Net~\cite{ma2020sfnet}  &53.8 &38.0 & 21.9&18.2 & 32.3   \\
            &  Our C$^3$BN  &55.1 &\bf{40.7} & \bf{22.9} & \bf{18.2} & \textbf{34.2}$_{\textcolor{blue}{+1.9}}$ \\
			\hline
			\multirow{5}{*}{BEOID}
			&DC~\cite{ju2021DC} & 63.2 & 46.8 & 20.9 & 5.8 & 34.9  \\
			\cline{2-7}
            &SF-Net~\cite{ma2020sfnet}  &60.3 &43.2 & 21.7 & 11.0 & 33.9   \\
            & Our C$^3$BN  &65.5 &44.0 & 26.3 & 9.7 & 37.0$_{\textcolor{blue}{+3.1}}$  \\
           \cline{2-7}
            &LACP~\cite{lee2021LACP}  &81.4 &73.1 &45.8 & 21.7 & 56.6  \\
            &  Our C$^3$BN  &\bf82.1 &\bf73.3 & \bf47.4 & \bf23.3 & \bf{57.6$_{\textcolor{blue}{+1.0}}$}  \\
			\hline
\end{tabular}
}
\end{center}
\end{table}

\subsection{Results on WTAL with Point-Level Labels}
There are also a few works proposed for WTAL with point-level supervision, \eg, SFNet~\cite{ma2020sfnet}, LACP~\cite{lee2021LACP}, and DC~\cite{ju2021DC}. Our C$^3$N is generic and is expected to work well for this task. We hereby take \textbf{SFNet}~\cite{ma2020sfnet} and \textbf{LACP}~\cite{lee2021LACP} as the baselines, and conduct experiments on three benchmark datasets: THUMOS14, BEOID, and GTEA.

We compare the proposed approach with recent methods for WTAL with point-level supervision in Table~\ref{table:comparison_small_datasets}. It can be seen that our C$^3$BN improves the performances of SF-Net~\cite{ma2020sfnet} and LACP~\cite{lee2021LACP} by a large margin. Besides, our method also outperforms the recently proposed DC~\cite{ju2021DC} and achieves SOTA performance on all three datasets. These results validate that our C$^3$BN is compatible with WTAL with different weak supervisions.

\begin{table}[t]
\begin{center}
		\begin{center}
			\setlength{\tabcolsep}{10pt}
			\caption{Results on ActivityNet v1.2. AVG indicates the average mAP at IoU thresholds 0.5:0.05:0.95.}
			\label{table:activity1.2}
			\resizebox{\columnwidth}{!}{
				\begin{tabular}{lcccl}
					\hline
					\cline{2-5}
					& 0.5 & 0.75 & 0.95 & AVG \\
					\hline
					TSCN~\cite{zhai2020two} & 37.6 & 23.7 & 5.7 & 23.6 \\
					WUM  \cite{lee2021weakly} & 41.2 & 25.6 & 6.0 & 25.9 \\
					CoLA \cite{CoLA} & 42.7 & 25.7 & 5.8 & 26.1 \\
					D2-Net \cite{narayan2021d2} & 42.3 & 25.5 & 5.8 & 26.0 \\
					ACGNet \cite{yang2021acgnet} & 41.8 & 26.0 & 5.9 & 26.1 \\
                        Li~\etal \cite{li2022forcing} & 41.6 & 24.8 & 5.4 & 25.2 \\
                        DELU~\cite{chen2022dual} & \bf{44.2} & 26.7 & 5.4 & 26.9 \\
					\hline
					FACNet \cite{huang2021foreground}  &41.2 &26.2 & 5.9 & 26.3 \\
					\rowcolor{gray!10} + Our C$^3$BN &43.9 &\bf{27.1} &\bf{6.3} &\bf{27.4$_{\textcolor{blue}{+1.1}}$} \\
					\hline
			\end{tabular}}
		\end{center}
		\hfil
\end{center}
\end{table}

\subsection{Results on ActivityNet v1.2}
In some previous WTAL methods (especially early methods), ActivityNet v1.2 is a preferred dataset rather than ActivityNet v1.3. Hence, we also evaluate our method on this benchmark. The results are presented in Table~\ref{table:activity1.2}. After combining with FACNet~\cite{huang2021foreground}, our method significantly outperforms previous methods, including the very recent method DELU (ECCV 2022). The favorable performances on all the benchmarks demonstrate the overall superiority of our proposed method.

\end{document}